\documentclass[a4paper]{article}

\usepackage[english]{babel}
\usepackage[utf8]{inputenc}
\usepackage[T1]{fontenc}
\usepackage[a4paper,top=2cm,bottom=2cm,left=2.5cm,right=2.5cm,marginparwidth=1.75cm]{geometry}

\usepackage{amsmath}
\usepackage{bm}
\usepackage{graphicx}
\usepackage{tabularx}
\usepackage{xcolor}
\definecolor{tcdBlue}{RGB}{5, 105, 185}
\usepackage[colorlinks=true,allcolors=.,urlcolor=blue]{hyperref} 

\usepackage{listings}

\usepackage[maxbibnames=5,style=ieee]{biblatex}

\def\BibTeX{{\rm B\kern-.05em{\sc i\kern-.025em b}\kern-.08em
    T\kern-.1667em\lower.7ex\hbox{E}\kern-.125emX}}

\usepackage{tabularray}
\UseTblrLibrary{booktabs}
\usepackage{threeparttable}
\usepackage{csquotes}

\usepackage{url}

\usepackage{titlesec}

\definecolor{eclipseStrings}{RGB}{42,0.0,255}
\definecolor{eclipseKeywords}{RGB}{127,0,85}
\colorlet{numb}{magenta!60!black}

\lstdefinelanguage{json}{
    basicstyle=\normalfont\ttfamily,
    commentstyle=\color{eclipseStrings}, 
    stringstyle=\color{eclipseKeywords}, 
    numbersep=8pt,
    showstringspaces=false,
    breaklines=true,
    frame=lines,
    string=[s]{"}{"},
    comment=[l]{:\ "},
    morecomment=[l]{:"},
    literate=
        *{0}{{{\color{numb}0}}}{1}
         {1}{{{\color{numb}1}}}{1}
         {2}{{{\color{numb}2}}}{1}
         {3}{{{\color{numb}3}}}{1}
         {4}{{{\color{numb}4}}}{1}
         {5}{{{\color{numb}5}}}{1}
         {6}{{{\color{numb}6}}}{1}
         {7}{{{\color{numb}7}}}{1}
         {8}{{{\color{numb}8}}}{1}
         {9}{{{\color{numb}9}}}{1}
}

\begin{document}

\title{\vspace{-1.5cm}Multimodal Structured Generation:\\CVPR's 2nd MMFM Challenge Technical Report}
\author{Franz Louis Cesista\\\textit{franzlouiscesista@gmail.com}}
\date{}
\maketitle

\begin{abstract}
    Multimodal Foundation Models (MMFMs) have demonstrated strong performance in both computer vision and natural language processing tasks. However, their performance diminishes in tasks that require a high degree of integration between these modalities, such as document understanding. Moreover, finetuning these models and deploying them requires significantly more compute and more engineering effort than unimodal models. In this work, we present \textbf{Multimodal Structured Generation}, a framework that forces (frozen) MMFMs to produce outputs in a strictly structured format by applying hard constraints directly to the output logits. This approach not only ensures that the model generates parseable outputs that downstream APIs can easily ingest but also allows us to force the model to \textit{reason} before answering, which significantly boosts performance without the need for expensive fine-tuning. We demonstrate the effectiveness of our method through competitive results in the CVPR 2nd MMFM Challenge, highlighting that carefully designed lightweight engineering can outperform expensive and complicated modeling approaches. \footnote{All of our scripts, deployment steps, and evaluation results can be accessed in \url{https://github.com/leloykun/MMFM-Challenge}}
\end{abstract}

\section{Introduction}

Multimodal Foundation Models (MMFMs) are typically built by grafting together components from different foundation models, each pretrained on a specific modality such as images or text \cite{bordes2024introduction}. The resulting "Frankenstein" models have shown impressive results on multimodal tasks where visual and textual data can be processed independently. For instance, in image captioning, a visual encoder first converts an image into a latent representation, which is then passed to a generative language model to produce a caption; similarly, in text-to-image generation, a language encoder transforms text input into a latent representation that an image generation model uses to create an image.

In contrast, document understanding requires the simultaneous processing and integration of visual, textual, and even layout information. And despite the progress made with MMFMs, their performance in tasks demanding deep multimodal integration remains suboptimal. Moreover, adapting these models for more specialized tasks via finetuning is both costly and labor-intensive.

To address these challenges, we introduce \textbf{Multimodal Structured Generation}, a framework that forces a (frozen) MMFM to generate outputs that strictly adhere to a predefined structure by applying hard constraints to its output logits. This mechanism allows us to:
\begin{enumerate}
    \item Force the model to generate outputs that follow the exact format downstream systems or APIs expect, thereby eliminating the need for bespoke post-processing steps. And,
    \item Force the model to \textit{reason} before producing an answer. This reduces the need for expensive fine-tuning, thereby reducing computational and engineering costs.
\end{enumerate}

As evidence of the practicality of our approach, we describe how we used this approach to win a prize in the Computer Vision and Pattern Recognition Conference's 2nd Multimodal Foundation Models Challenge. Our team learned about the challenge only two days before the submissions deadline, and we spent the first 24 hours working with a commercially available model that, after clarification with the organizers, we were not permitted to use. With limited time, compute resources, and budget, we could not afford to implement complicated modeling steps. Thus, we turned to our prior work on \textit{Retrieval Augmented Structured Generation} \cite{cesista2024rasg}.

Retrieval Augmented Structured Generation has four components: (1) Structured Generation \cite{willard2023outlines}, (2) Retrieval Augmented Generation \cite{ram2023incontext}, (3) Supervised Finetuning, and (4) Structured Prompting \cite{wang2023latin}. However, for this challenge, we only implemented the Structured Generation component, extending it to multimodal models instead of solely on a language model as in our previous work.

Our approach proved effective: we placed 2nd in Phase 2 of CVPR's 2nd MMFM Challenge and 3rd overall, outperforming several teams that had finetuned their own multimodal models. Notably, Phase 2 featured various never-before-seen document understanding evaluation datasets, demonstrating the generality of our method to unseen tasks. Furthermore, the fact that our method is finetuning-free makes it both cost-effective and easy for other teams to replicate.

\section{Methodology}

\begin{figure}[t]
    \centering
    \includegraphics[width=0.6\linewidth]{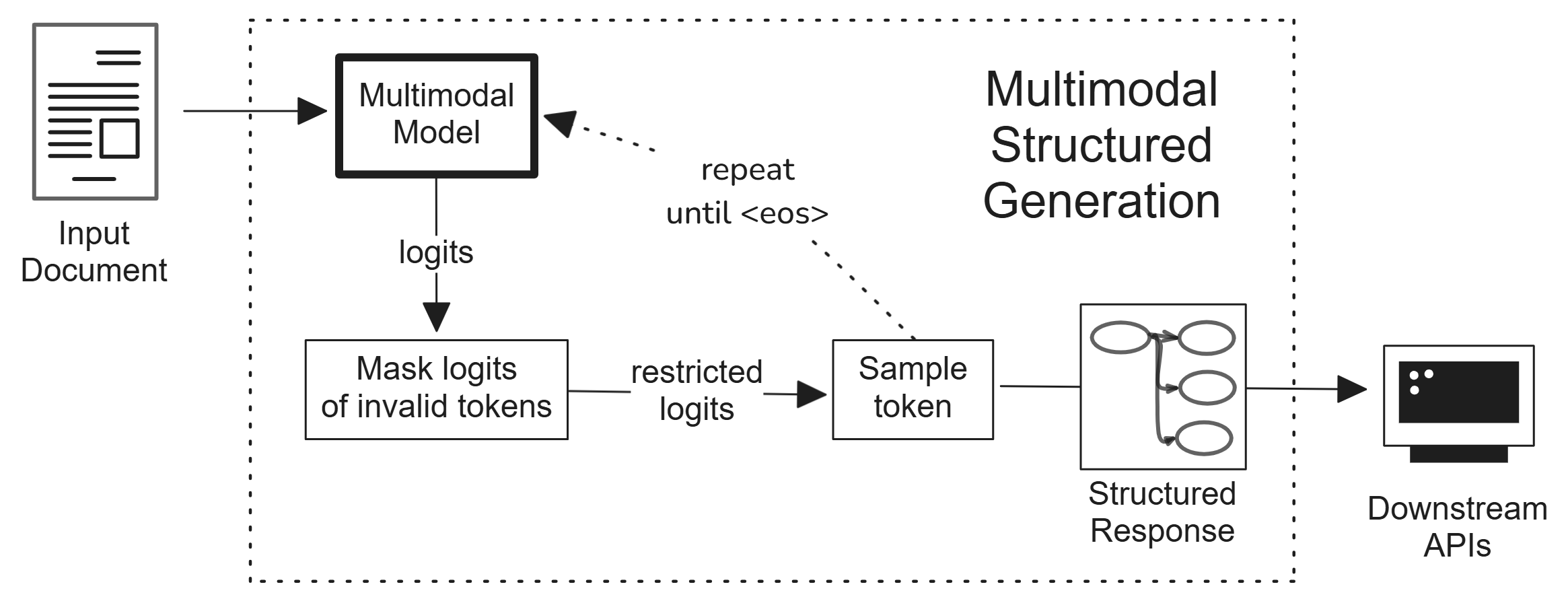}
    \caption{Multimodal Structured Generation}
    \label{fig:multimodal-structured-gen}
\end{figure}

\subsection{Multimodal Structured Generation}

Outputs from generative models are not guaranteed to be parseable by downstream programs, compilers, or APIs. For example, if a user requests that a language model generate a Python script for sorting a list of numbers, the output might \textit{seem} runnable yet produce errors when executed in an IDE. To ensure that generated outputs are immediately usable, our approach enforces hard constraints that zero out the logits corresponding to invalid tokens \cite{willard2023outlines}. This method, known as Structured Generation, is readily applicable to multimodal models, which also produce logits that can be constrained to adhere strictly to a predefined schema (see Figure \ref{fig:multimodal-structured-gen}).

\subsection{Implementation Details}

We deploy our multimodal models using Huggingface's Inference Endpoints API and perform inference via Huggingface's Text Generation Interface (TGI) API \cite{huggingface2024inference} \cite{huggingface2024tgi}.

For Phase 1, since the test dataset was publicly available and did not fully reflect the generality of our approach, we used an unmodified Llava v1.5 \cite{liu2023llava}. For Phase 2, which featured never-before-seen evaluation datasets, we augmented our framework with two models. For the \texttt{mychart} and \texttt{myinfographic} datasets, we employed Llava-Next (v1.6) enhanced with Structured Generation \cite{liu2024llavanext} \cite{liu2023improvedllava}. And to maximize performance on the \texttt{mydoc} dataset, we used a finetuned version of Nous Hermes 2 Pro - Mistral 7B\footnote{This model can be accessed in \url{https://huggingface.co/leloy/Nous-Hermes-2-Pro-Docile-RASG-1ShotRetrieval-StructuredPrompt}} augmented with Structured Generation \cite{anon2024hermespro}.

Our Structured Generation method forces the multimodal models to "reason" before producing an answer. We designed slightly different JSON output formats for each evaluation dataset (available in our repository). For the \texttt{myinfographic} and \texttt{mychart} datasets, a template can be found in Appendix \ref{appendix:json-template-1}. For the \texttt{mydoc} dataset, we used the entity being requested as the key, following the JSON format from our previous work \cite{cesista2024rasg}. This design makes it straightforward to request multiple entities from a document (e.g., \texttt{Billing Name} and \texttt{Total Amount}). A template can be found in Appendix \ref{appendix:json-template-2}.

Furthermore, we instructed the models to provide exact answers for the \texttt{mydoc} dataset while requesting concise answers for the \texttt{mychart} and \texttt{myinfographic} datasets. We suspected that the challenge organizers used the MMMU evaluation metric for \texttt{mydoc} and judged the outputs for \texttt{mychart} and \texttt{myinfographic} with the Mistral 7B model as LLM-as-judge \cite{yue2023mmmu} \cite{mistral2024mistral7b}. The former requires exact outputs, whereas the evaluation script for the latter indicates that concise responses are preferred. We later learned after the challenge that the Mistral 7B model was used as the judge across all datasets. However, we believe that the exact answers for the \texttt{mydoc} dataset contributed positively to our final results.

\subsection{Datasets}

The MMFM challenge provided the following datasets for Phase 1: IconQA \cite{lu2022iconqa}, FUNSD \cite{jaume2019funsd}, WildReceipt \cite{sun2021wildreceipt}, TextbookQA \cite{kim2019textbookqa}, TabFact \cite{chen2020tabfact}, DocVQA \cite{mathew2021docvqa}, InfographicVQA \cite{mathew2021infographicvqa}, WebSRC \cite{chen2021websrc}, and WTQ \cite{drozdov2022compositional}.

\section{Results}

We evaluated our approach on the two phases of CVPR's 2nd MMFM's Challenge. See Table \ref{tab:results} for results. Our approach placed 2nd in the hidden test set of Phase 2 and 3rd place overall.

{
\begin{table}[h]
\centering
\caption{Evaluation results on CVPR's 2nd MMFM Challenge}
\begin{tabular}{ccr}
    & \textbf{Eval Dataset} & \textbf{Accuracy} \\
    \midrule \midrule
    Phase 1 & \texttt{iconqa\_fill} & $15.5\%$ \\
    & \texttt{funsd} & $32.5\%$ \\
    & \texttt{iconqa\_choose} & $31.0\%$ \\
    & \texttt{wildreceipt} & $35.5\%$ \\
    & \texttt{textbookqa} & $51.5\%$ \\
    & \texttt{tabfact} & $48.5\%$ \\
    & \texttt{docvqa} & $20.5\%$ \\
    & \texttt{infographicvqa} & $23\%$ \\
    & \texttt{websrc} & $28.5\%$ \\
    & \texttt{wtq} & $8.5\%$ \\
    & \textbf{Phase 1 Overall} & $\bm{29.5\%}$ \\
    \midrule
    Phase 2 & \texttt{mydoc} & $62.25\%$ \\
    & \texttt{mychart} & $4.5\%$ \\
    & \texttt{myinfographic} & $60.98\%$ \\
    & \textbf{Phase 2 Overall} & $\bm{50.49\%}$ \\
\end{tabular}
\label{tab:results}
\end{table}
}

\section{Discussion}

Our results on the \textsc{mydoc} dataset are particularly noteworthy: we outperformed multiple teams that finetuned multimodal (vision + text) models by using only an LLM augmented with Structured Generation. This finding reinforces a key insight from our previous work \cite{cesista2024rasg}—namely, that for key-information extraction, visual information may be less critical than expected, and in some cases, incorporating vision encoders can even be detrimental.

{
\begin{table}[t]
\centering
\caption{Ablation Benchmarks of RASG components on KIE \& LIR tasks on the DocILE dataset}
\begin{threeparttable}
\begin{tabular}{lrr}
    \toprule
    \textbf{Model} & \textbf{Key-Information Extraction} & \textbf{Line Items Recognition} \\
    & \textbf{F1 Score} & \textbf{GLIRM-F1} \\
    \midrule
    \textbf{GPT-3.5}        &   $34.17\%$ &   $28.31\%$ \\
    + 1-Shot Retrieval      & + $22.08\%$ & + $20.67\%$ \\
    + Supervised Finetuning & + $22.31\%$ & + $17.73\%$ \\
    + Structured Prompting  & + $4.96\%$  & + $19.42\%$ \\
    \midrule
    \textbf{Hermes 2 Pro - Mistral 7B} & $13.55\%$ & $4.69\%$ \\
    + 1-Shot Retrieval      & + $36.87\%$ & + $40.55\%$ \\
    + Supervised Finetuning & + $17.71\%$ & + $13.53\%$ \\
    + Structured Prompting  & +  $0.63\%$ & + $10.30\%$ \\
    \bottomrule
\end{tabular}
\begin{tablenotes}
    \item[*]Benchmarks results ablating three components of Retrieval Augmented Structured Generation on Key-Information Extraction (KIE) \& Line Items Recognition (LIR) tasks on the DocILE dataset \cite{simsa2023docile}: (1) Retrieval Augmented Generation, (2) Supervised Finetuning, and (3) Structured Prompting. Structured Generation was not included in the ablation benchmarks as it is a necessary component of RASG to ensure that the outputs are parse-able by downstream APIs. Results show that adding Structured Prompting, i.e., adding layout information to the text prompt, only marginally improves performance.
\end{tablenotes}
\end{threeparttable}
\label{tab:rasg-table}
\end{table}
}

We propose four hypotheses to explain this observation:

\begin{enumerate}
    \item The visual and layout information may not be essential for Key-Information Extraction.

    In support of this observation, DocLLM demonstrated that a Text + Layout model can perform as well as, or even outperform, models that integrate both visual and layout cues \cite{wang2023docllm}. Furthermore, our previous work showed that, with appropriate augmentations, entirely omitting layout information does not significantly degrade performance \cite{cesista2024rasg}. This finding suggests that the critical factor for successful KIE is the preservation of \textit{locality}—ensuring that words appearing together in the document remain proximate in the text prompt—a property that is naturally maintained by accurate OCR outputs.

    \item The LLMs themselves may be capable of inferring the spatial arrangement of words from their sequential positions in the text prompt.
    
    Prior work has shown that LLMs trained without explicit positional encodings can still learn and exploit positional information \cite{haviv2022transformer}. It is plausible that, when the text is provided in the order extracted by OCR, the LLM implicitly reconstructs the spatial relationships among words.

    \item The limited information capacity of the base language models may also play a role.

    For example, the LLaVA-NeXT (v1.6) model we used is based on the Mistral-7B LLM \cite{liu2024llavanext}. A rough estimation using neural scaling laws \cite{kaplan2020scaling} suggests that such models can store roughly \(7 \times 20 = 140\) billion tokens of information. When additional embeddings from the visual encoder and projector are incorporated, the model may quickly reach its capacity limits, thereby reducing its effectiveness for complex document understanding tasks.

    \item Perhaps most compellingly, we may not be using enough image tokens for effective document understanding.

    Unlike images of pets, which can be adequately described with just a few tokens, document images are dense with information. Empirically, business documents often contain around 3,000 text tokens—and this number increases when layout details are considered. However, standard image encoders typically compress an entire document into no more than $32^2 = 1024$ tokens, leading to significant information loss. Recent work \cite{cai2024matryoshka} supports this view, showing that increasing the number of image tokens improves performance on document understanding benchmarks (see Figure \ref{fig:matryoshka-mm}).

    \begin{figure}
        \centering
        \includegraphics[width=0.9\linewidth]{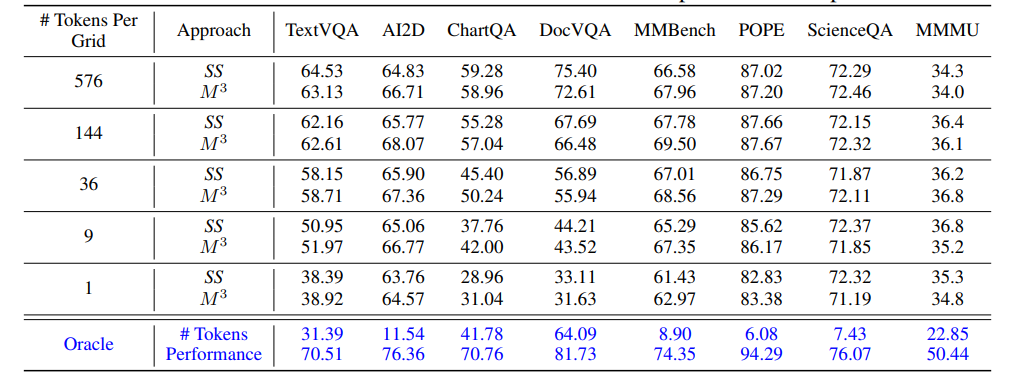}
        \caption{Comparison of approaches with the SS baseline and Matryoshka Multimodal Models (M${^3}$) across various benchmarks under LLaVA-NeXT \cite{cai2024matryoshka}. The number of tokens (\# Tokens) denotes the number of visual tokens per image grid in LLaVA-NeXT. SS denotes the baseline model trained with a Specific Scale of visual tokens. M${^3}$ is at least as good as SS, while performing better on tasks such as TextVQA, ChartQA, and MMBench. Oracle denotes the case where the best tradeoff between visual tokens and performance is picked.}
        \label{fig:matryoshka-mm}
    \end{figure}
\end{enumerate}

\section{Conclusion}

Our final standings in CVPR's 2nd MMFM Challenge demonstrate that our approach generalizes effectively to unseen tasks. Notably, our finetuning-free implementation not only reduces computational and engineering overhead, but also makes the method readily reproducible for other teams. While our framework can be extended to finetuned models, our previous work indicates that naively combining finetuning with Structured Generation may actually degrade performance \cite{cesista2024rasg}. Future research should further explore the optimal integration of these components, with particular attention to maintaining alignment between output formats in both the finetuning dataset and inference requests.

Additionally, our results on the \texttt{mydoc} dataset reinforce the claim that neither detailed visual cues nor precise layout information is critical for effective key-information extraction. Nevertheless, further investigation is needed to rigorously evaluate the four hypotheses proposed in the discussion. Overall, our findings underscore the promise of lightweight engineering approaches in advancing document understanding tasks while reducing reliance on expensive, complex modeling techniques.

\break

\printbibliography

\break

\appendix

\section{JSON Output Formats}

\subsection{JSON output format for the \texttt{myinfographic} and \texttt{mychart} datasets}\label{appendix:json-template-1}

\begin{lstlisting}[language=json]
{
  "name": "<tool name e.g. infographic_explair_tool>",
  "description": "<tool description e.g. Infographic Explainer Tool>",
  "parameters": {
    "type": "object",
    "properties": {
      "1_reasoning": {"type": "string"},
      "2_answer": {
        "type": "string",
        "description": "Concise answer to the user question."
      },
    },
    "required": ["1_reasoning", "2_answer"],
  },
}
\end{lstlisting}

\subsection{JSON output format for the \texttt{mydoc} datasets}\label{appendix:json-template-2}

\begin{lstlisting}[language=json]
{
  "name": "doc_extraction_tool",
  "description": "Extract information from a document",
  "parameters": {
    "type": "object",
    "properties": {
      "1_reasoning": {"type": "string"},
      f"2_{key}": {
        "type": "integer" if key == "page" else "string",
        "description": "The answer, exactly as it appears in the document.",
        "maxLength": max_length,
      }
    },
    "required": ["1_reasoning", f"2_{key}"],
  },
}
\end{lstlisting}

Note that we prepend indices to the keys in the JSON format because the version of TGI we used relies on an older version of Outlines (version $< 0.40.0$) that reorders keys alphabetically. Prepending indices is an effective workaround to ensure the desired order; these indices can be removed in later versions of TGI and Outlines.

\end{document}